\documentclass[aps,12pt,epsf]{revtex4}

\usepackage{amsmath}
\usepackage{amssymb}
\usepackage{graphicx}
\usepackage{comment}
\usepackage{bbold}

\begin{document}

\title{Reward-estimation variance elimination in sequential decision processes}

\author{Sergey \surname{Pankov}}
\affiliation{Harik Shazeer Labs, Palo Alto, CA 94301}

\begin{abstract}

Policy gradient methods are very attractive in reinforcement learning due to their model-free nature and convergence guarantees. These methods, however, suffer from high variance in gradient estimation, resulting in poor sample efficiency. To mitigate this issue, a number of variance-reduction approaches have been proposed. Unfortunately, in the challenging problems with delayed rewards, these approaches either bring a relatively modest improvement or do reduce variance at expense of introducing a bias and undermining convergence. The unbiased methods of gradient estimation, in general, only partially reduce variance, without eliminating it completely even in the limit of exact knowledge of the value functions and problem dynamics, as one might have wished. 

In this work we propose an unbiased method that does completely eliminate variance under some, commonly encountered, conditions. Of practical interest is the limit of deterministic dynamics and small policy stochasticity. In the case of a quadratic value function, as in linear quadratic Gaussian models, the policy randomness need not be small. We use such a model to analyze performance of the proposed variance-elimination approach and compare it with standard variance-reduction methods.

The core idea behind the approach is to use control variates at all future times down the trajectory. We present both a model-based and model-free formulations.

\end{abstract}

\maketitle

\newcommand{\T}{\mathsf T}

\section{introduction}

Deep reinforcement learning -- reinforcement learning (RL) with expressive function approximators -- has been successful in tackling challenging RL problems \cite{mnih2013playing,schulman2015trust,silver2016mastering,gu2017deep}. In many cases, impressive results were obtained with the algorithms actually lacking converges guarantees, such as various variants of Q-learning \cite{riedmiller2005neural,mnih2013playing,lillicrap2015continuous}. This lack of convergence can turn into a significant obstacle when scaling these algorithms up to more complex problems, with higher dimensions, longer reward delays or intrinsically more complex policies.

Policy gradient (PG) methods \cite{aleksandrov1968stochastic,sutton2000policy,peters2006policy,kakade2002natural,schulman2015trust}, a class of RL algorithms, do offer convergence guaranties in principle, as they directly optimize the expected reward via gradient ascent. In practice, it often comes at the price of a large gradient sampling variance, leading to poor sample efficiency. This may hinder the application of PG methods to real world problems, when collecting trajectory samples is not cheap. This is a general problem, also known as the credit assignment problem \cite{sutton1984temporal}. When many actions are taken before a reward (credit) arrives, it can be difficult to discern which actions contributed positively or negatively to the reward. It thus may take many samples to filter out statistical noise. This can be especially acute in problems with fine-grained dynamics and complex policies, for example, in the continuous-time underactuated control tasks.

To ameliorate this issue, various variance reduction methods have been proposed \cite{williams1992simple,sutton2000policy,schulman2015high}. They broadly fall in to two groups, with unbiased and biased variance reduction. One popular unbiased approach is to subtract a (state-dependent) baseline \cite{sutton2000policy} from the accumulated reward. This procedure does not affect the gradient expectation, thus reducing variance without any adverse effects\cite{williams1992simple,sutton2000policy}. But the remaining variance can still be large in problems with delayed rewards and long horizons. The biased approaches tend to reduce the variance more aggressively, but at the price of introducing bias \cite{schulman2015high}. While they can be useful in practice, navigating a delicate trade-off between the variance reduction and biased gradient estimation may require extensive problem-dependent hyperparameter tuning.

More recently state-action-dependent baselines have been investigated \cite{gu2016q,grathwohl2017backpropagation,liu2018action,wu2018variance}. The general idea is to include more information into the baseline to be able to better counter fluctuations of the sampled quantity. Unlike state-dependent baselines, the action-dependent corrections do not vanish in the gradient, so they must be appropriately compensated. For this reason they should be of a specific form, permitting efficient expectation evaluation, for example, by means of analytic integration. This is known as the method of control variates.

However, a careful analysis of the action-dependent control variate methods revealed their relatively small contribution to variance reduction \cite{tucker2018mirage}. Indeed, it is natural to expect that the actions following a particular action, unless heavily discounted, would each contribute significantly to the variance. Therefore a significant variance reduction requires taking care of the states and actions down the trajectory, not just the current state-action pair.

We present such an approach in this paper. We apply the control variates method at all future times, in a way that reduces variance without introducing a bias. Moreover, in the limit of deterministic dynamics and quadratic approximation to the value function (small policy stochasticity, in other words), the variance is completely eliminated. (We therefore call it the variance elimination (VE) method.) This limit is relevant to many practical problems, with robotic locomotion being one example. The approach is still valid when the above conditions are not met. But in that case only partial variance reduction is possible.

The paper is organized as follows. In section \ref{unbve} we derive VE, both in model-based and model-free frameworks. In section \ref{analysis} we compare performances of VE and other standard baseline methods by applying them to a controlled diffusion process model. 
In section \ref{discussion} we discuss possible VE-related developments.

\section{Unbiased variance elimination}
\label{unbve}

Consider a Markov decision process $\{\mathcal{S,A,T},r,\gamma,N,s_0\}$. At time $t$: 1) a state $s_t\in\mathcal{S}$ is sampled (unless $t=0$) according to $p(s_t|s_{t-1},a_{t-1})$ defined by the transition probability tensor $\mathcal{T}$, 2) an action $a_t\in\mathcal{A}$ is sampled according to the policy $\pi(a_t|s_t)$, 3) a reward $r_t = r(s_t,a_t)$, defined by $r:\mathcal{S\times A}\to\mathbb{R}$, is issued. The process is repeated until the horizon $t=N$, which can be finite or infinite. The discount factor $0<\gamma\le 1$ is used for discounting later rewards when computing a cumulative reward. In the undiscounted reward case $\gamma=1$ we assume a finite horizon $N$.

The goal of RL is to find a policy that maximizes the expected reward $\mathbb{E}_{\tau\sim\pi,p}[\sum_{t=0}^{t=N} \gamma^t r_t]$, computed over the trajectories $\tau$ sampled under the policy $\pi$ and dynamics $p$. We define the state, state-action and state-action-state value functions:
\begin{equation}
\begin{split}
& V(s_t) = E_{\tau\sim\pi,p}\left[ \sum_{i=t}^{i=N} \gamma^{i-t} r_i | s_t \right] \\
& Q(s_t,a_t) = E_{\tau\sim\pi,p}\left[ \sum_{i=t}^{i=N} \gamma^{i-t} r_i | s_t,a_t \right] \\
& R(s_t,a_t,s_{t+1}) = E_{\tau\sim\pi,p}\left[ \sum_{i=t}^{i=N} \gamma^{i-t} r_i | s_t,a_t,s_{t+1} \right]
\end{split}
\end{equation}
Notice that $V(s)=\mathbb{E}_{a\sim\pi(a|s)}[Q(s,a)]$ and $Q(s,a)=\mathbb{E}_{s'\sim p(s'|s,a)}[R(s,a,s')]$.

We will use the method of control variates -- a method for reducing variance of an expectation estimate. Consider the problem of estimating $E[f(x)]$ by sampling $x$. Notice that for any $g(x)$ we have $\mathbb{E}[f(x)]=\mathbb{E}[f(x)-g(x)]+\mathbb{E}[g(x)]$. One aims at choosing $g(x)$ that: a) has a known expectation value $\mathbb{E}[g(x)]$, so only the first term needs to be estimated by sampling, b) is close to $f(x)$, so the variance of the estimate is small.

We provide both a model-based and model free formulations. In the model-based case, the reward model $r$, the state value function $V$ and the transition model $p$ are approximated. This approach may be suitable if the dynamics is known. It could be known exactly, or could have been learned solving a different problem (it is not uncommon to consider multiple problems with different reward setups but shared dynamics). In the model-free case only the state-action value function $Q$ needs to be approximated and this formulation appears conceptually simpler. But even when the dynamics are unknown, it is not a priori obvious whether treating $r$, $V$ and $p$ implicitly via a single function $Q$ should generally be a preferred choice.

\subsection{Model-based formulation}

Let us first consider a model-based approach. We use a tilde to denote main approximators -- the approximate quantities that are used to define all other quantities in the following derivation. These include the reward function $\tilde r(s,a)$, the state value function $\tilde V(s)$ and the dynamics $\tilde p(s'|s,a)$ in the stochastic case or $\tilde f(s,a)$ in the deterministic case. We use an overbar to indicate approximate quantities that are expressed in a specific and (computationally) straightforward way via the main approximators. These include
\begin{equation}
\begin{split}
& \bar R(s,a,s') = \tilde r(s,a) + \gamma\tilde V(s') \\
& \bar Q(s,a)=\mathbb{E}_{s'\sim \tilde p(s'|s,a)}[\bar R(s,a,s')] \\
& \bar V(s)=\mathbb{E}_{a\sim\pi(a|s)}[\bar Q(s,a)]
\end{split}
\label{rqvbars}
\end{equation}
Notice that in general $\bar V(s) \ne \tilde V(s)$. Only if the main approximators are exact, do we have $\bar V(s)=\tilde V(s)=V(s)$. Our general approach is to add and subtract an expectation of some quantity, evaluating the subtracted part by sampling. Let us first consider a somewhat academic case of $\tilde p = p$, which is unlikely to arise in a physical world problem, but which demonstrates the idea in a general stochastic setting. Adding and subtracting $\mathbb{E}_{a\sim\pi(a|s_i), s'\sim \tilde p(s'|s_i,a)}[\bar R(s_i,a,s')]$ to $r_i$, we get:
\begin{equation}
 r_i \to r_i - \tilde r(s_i,a_i) - \gamma\tilde V(s_{i+1}) + \mathbb{E}_{\substack{a\sim\pi(a|s_i) \\ s'\sim \tilde p(s'|s_i,a)}}[\bar R(s_i,a,s')]
\end{equation}
Notice that we used $(a_i,s_{i+1})$ as a sample drawn from the expectation distribution (because $\tilde p = p$), and the expectation term is $\bar V(s_i)$. 
Repeating this for every $r_i$ in the discounted reward sum $\sum_{i=t}^{i=N} \gamma^{i-t} r_i$ we obtain an expected cumulative reward estimator $\hat V_t$:
\begin{equation}
  \hat V_t = \bar V(s_t) 
+ \sum_{i=t}^{i=N} \gamma^{i-t}(r_i - \tilde r(s_i,a_i))
+ \sum_{i=t+1}^{i=N} \gamma^{i-t} \left( \bar V(s_i) - \tilde V(s_i) \right)
\end{equation}
By construction, $\hat V_t$ is an unbiased estimator of $V(s_t)$. Moreover, in the limit of the exact approximation (when $\tilde r = r$ and $\tilde V = \bar V = V$), the variance of the estimator is zero. However, for this result to become practically useful, one needs to be able to efficiently compute $\bar V(s)$, in addition to requiring $\tilde p = p$.

Consider now $\tilde p \ne p$. In this case the control variates method is only applicable to the sampling from $\pi$, so we add and subtract $\mathbb{E}_{a\sim\pi(a|s_i)}[\bar Q(s_i,a)]$ to $r_i$. Note that in expectation this is $\bar V(s)$, just as before. 
For the state value estimator we get:
\begin{equation}
  \hat V_t = \bar V(s_t) 
+ \sum_{i=t}^{i=N} \gamma^{i-t}(r_i - \tilde r(s_i,a_i))
+ \sum_{i=t+1}^{i=N} \gamma^{i-t} \left( \bar V(s_i) - \mathbb{E}_{s'\sim \tilde p(s'|s_{i-1},a_{i-1})}[\tilde V(s')] \right)
\end{equation}
The above expression is still an unbiased estimator, but its variance may remain finite in the limit of the exact approximation, because the last sum does not vanish, in general, as before. It only vanishes in the limit of deterministic dynamics. But this is a practically important case, as in many interesting problems the dynamics is either deterministic, or only weakly stochastic, in which case one can still expect a substantial variance reduction. In the deterministic dynamics case we write:
\begin{equation}
  \hat V_t = \bar V(s_t) 
+ \sum_{i=t}^{i=N} \gamma^{i-t}(r_i - \tilde r(s_i,a_i))
+ \sum_{i=t+1}^{i=N} \gamma^{i-t} \left( \bar V(s_i) - \tilde V(\tilde s_i) \right)
\label{mbest}
\end{equation}
where we have defined $\tilde s_{t+1} = \tilde f(s_t,a_t)$.

\subsection{Model-free formulation}

In the model-free formulation we define $\tilde Q$ as the main approximator, instead of $\tilde r, \tilde V$ and $\tilde f$. We then define $\bar V$:
\begin{equation}
\bar V(s)=\mathbb{E}_{a\sim\pi(a|s)}[\tilde Q(s,a)]
\label{vbarmf}
\end{equation}
Adding and subtracting $\mathbb{E}_{a\sim\pi(a|s_i)}[\tilde Q(s_i,a)]$ to every $r_i$, after some re-arranging we get:
\begin{equation}
  \hat V_t = \bar V(s_t)
+ \sum_{i=t}^{i=N-1} \gamma^{i-t} \left( r_i + \gamma\bar V(s_{i+1}) - \tilde Q(s_i,a_i) \right)
+\gamma^{N-t}\left( r_N - \tilde Q(s_N,a_N)\right)
\label{mfvest}
\end{equation}
In the case of deterministic dynamics, the above sum can be recognized as a discounted sum of temporal differences. Again, in the limit of exact approximation the last two terms vanish, making the variance zero. And again, in the case of stochastic dynamics, the variance is in general non-zero. Equivalence of Eq.(\ref{mbest}) and Eq.(\ref{mfvest}) can be established by substituting $\tilde Q(s,a)$ with $\tilde r(s,a) + \gamma\tilde V(\tilde s')$.

The construction of the state-action value estimator $\hat Q_t$ proceeds almost identically to $\hat V_t$, except for $r_t$ we add and subtract $\tilde Q(s_t,a_t)$ to obtain:
\begin{equation}
  \hat Q_t = \tilde Q(s_t,a_t)
+ \sum_{i=t}^{i=N-1} \gamma^{i-t} \left( r_i + \gamma\bar V(s_{i+1}) - \tilde Q(s_i,a_i) \right)
+\gamma^{N-t}\left( r_N - \tilde Q(s_N,a_N)\right)
\label{mfqest}
\end{equation}
It is easy to see that the estimator $\hat Q_t$ obeys a simple recursive relation:
\begin{equation}
  \hat Q_{t-1} = r_{t-1} + \gamma\bar V(s_t) + \gamma\left( \hat Q_t - \tilde Q(s_t,a_t) \right)
\label{qestrecur}
\end{equation}
with $\hat Q_N = r_N$.

When computing a PG one faces the problem of estimating by sampling $\mathbb{E}_{a\sim\pi(a|s)}[\nabla\ln{\pi(a|s)} \hat Q]$, with the gradient taken with respect to the policy parameters. Using again the control variates method we evaluate instead
\begin{equation}
\mathbb{E}_{a\sim\pi(a|s)}\left[\nabla\ln{\pi(a|s)} \left(\hat Q - \tilde Q(s,a)\right)\right] + \nabla\bar V(s)
\label{vegrad}
\end{equation}
where $\nabla\bar V(s) = \mathbb{E}_{a\sim\pi(a|s)}[\nabla\ln{\pi(a|s)} \tilde Q(s,a)]$. The estimation variance vanishes under the same conditions as for the value functions $V$ and $Q$.

\subsection{Design of approximators}

The main approximators can be designed in many ways. For example, they can be based on some theoretical considerations, as in \ref{comparison} . Or they can be derived from empirical data, as we explain below. Whichever way is chosen, it must permit efficient computation of $\bar V$. 

 We assume continuous states and actions, and a Gaussian policy $\pi(a|s)$, that is $a\sim\mathcal{N}(\bar a(s), W)$. Let us consider the model-based case first. We need functions $r'(s,a)$, $V'(s,a)$ and $f'(s,a)$ that approximate $r(s,a)$, $V(s,a)$ and $f(s,a)$ respectively, and are appropriately differentiable (see below). One can assume that $r'$, $V'$ and $f'$ are neural-net function approximators trained on $(s,a;r)$, $(s;\hat V)$ and $(s,a;s')$ respectively. We consider a quadratic approximation for $\tilde r$ and $\tilde V$ by expanding $r'$ and $V'$ to second order around $\bar a = \bar a(s)$ and $\bar s = \tilde f(s,\bar a)$ respectively.  We consider a linear approximation for $\tilde f$ by expanding $f'$ to first order around $\bar a$. It is straightforward to include higher order terms as well. Writing $\tilde f$, $\tilde r$, $\tilde V$ and $\bar V$ (defined in Eq.(\ref{rqvbars})) explicitly:
\begin{equation}
\begin{split}
& \tilde f(s,a) = f'^{(0)} + f'^{(1)} (a-\bar a) \\
& \tilde r(s,a) = r'^{(0)} + r'^{(1)} (a-\bar a) + \frac{1}{2} (a-\bar a)^\T r'^{(2)} (a-\bar a) \\
& \tilde V(s) = V'^{(0)} + V'^{(1)} (s-\bar s) + \frac{1}{2} (s-\bar s)^\T V'^{(2)} (s-\bar s) \\
& \bar V(s) = r'^{(0)} + \gamma V'^{(0)} +
\frac{1}{2} \textrm{Tr}\left\{\left(r'^{(2)} + \gamma{f'^{(1)}}^\T V'^{(2)} f'^{(1)} 
\right)W\right\}
\end{split}
\end{equation}
where we have defined the derivatives:
\begin{equation}
f'^{(n)} = \frac{\partial^n f'(s,\bar a)}{\partial \bar a^n}, \qquad
r'^{(n)} = \frac{\partial^n r'(s,\bar a)}{\partial \bar a^n}, \qquad
V'^{(n)} = \frac{\partial^n V'(\bar s)}{\partial \bar s^n}.
\end{equation}
If $f$ is linear, $r$ (and hence $V$) is quadratic, $f'\to f$, $r'\to r$ and $V'\to V$, then the approximators become exact, that is $\tilde f \to f$, $\tilde r \to r$, $\tilde V \to V$ and $\bar V \to V$. If $\tilde f$ is considered to the second order, $f'\to f$, $r'\to r$ and $V'\to V$, then we have $\bar V = V +\mathcal{O}(W^2)$ and complete variance elimination in the limit of small policy stochasticity $W\to 0$.

Let us consider the model-free case now. Consider a function $Q'(s,a)$ that approximates $Q(s,a)$ and is twice differentiable with respect to $a$. We construct the approximator $\tilde Q(s,a)$ by expanding $Q'$ to second order around $\bar a$. Writing $\tilde Q$ and $\bar V$ (defined in Eq.(\ref{vbarmf})) explicitly:
\begin{equation}
\begin{split}
& \tilde Q(s,a) = Q'^{(0)} + Q'^{(1)} (a-\bar a) + \frac{1}{2}(a-\bar a)^\T Q'^{(2)} (a-\bar a) \\
& \bar V(s) = Q'^{(0)} + \frac{1}{2}\textrm{Tr}\left\{Q'^{(2)}W\right\}
\end{split}
\end{equation}
where $Q'^{(n)} = {\partial^n Q'(s,\bar a)}/{\partial \bar a^n}$. Similarly to the model-based case, if $Q$ is quadratic, and $Q'\to Q$, then the approximators become exact, $\tilde Q \to Q$ and $\bar V \to V$. If $Q'\to Q$, then we have $\tilde Q = Q +\mathcal{O}(W^2)$, $\bar V = V +\mathcal{O}(W^2)$ and complete variance elimination for $W \to 0$. When we talk about small policy stochasticity, we imply that $\textrm{Tr}\left\{Q'^{(2)}W\right\}$ is much larger than the higher order terms in the expansion of $V$ in powers of $W$, denoted as $\mathcal{O}(W^2)$.

\section{LQG analysis of the variance reduction and elimination}
\label{analysis}

In this section we demonstrate performance of VE on a controlled diffusion model \cite{krylov2008controlled} -- a simple linear quadratic Gaussian (LQG) model \cite{stengel1994optimal}. In this model the value functions are readily computed exactly, in the continuous time limit. We employ them to construct approximators for VE. We then compare VE to other variance reduction methods, that use a constant, state dependent and state-action dependent baselines.  

\subsection{Controlled diffusion model}
\label{contrdiff}
 
We consider the LQG model with very simple dynamics, describing one dimensional diffusion of massless particles. In the discrete time formulation the scalar state $s$ is governed by the linear dynamics:
\begin{equation}
s' = s + B^d \bar a + \eta
\label{diffdyn}
\end{equation}
where $\eta$ is a random Gaussian variable $\eta \sim \mathcal{N}(0,W^d)$. We will treat $\eta$ as a stochastic contribution to the action $a=\bar a + \eta/B^d$, where $\bar a$ is the deterministic part. The reward function is quadratic $r(s,a) = -C_s^d s^2 - C_a^d a^2$. The continuous time formulation is recovered by sending the time step $\Delta\to0$, while scaling $B^d = \Delta B$, $W^d = \Delta W$, $C_s^d=\Delta C_s$ and $C_a^d=\Delta C_a$ so that $B$, $W$, $C_s$ and $C_a$ are fixed. In the remainder of the paper we assume undiscounted rewards ($\gamma=1$) and a finite horizon $N$. We also define $T=(N+1)\Delta$. Note that the $i$-th time step in discrete time formulation corresponds to $t=i\Delta$ in continuous time formulation.

It is known that the optimal control law for LQG has the linear form $\bar a = -K (s-\mu_\infty)$ \cite{stengel1994optimal}, (with appropriately chosen $K$ and $\mu_\infty$). Assume that the initial state distribution is Gaussian, that is $s_0 \sim \mathcal{N}(\mu,\Sigma)$. Under the diffusion dynamics Eq.(\ref{diffdyn}) the distribution will remain Gaussian at all times, with the parameters $\mu(t)$ and $\Sigma(t)$, in the continuous dynamics limit, given by 
\begin{equation}
\begin{split}
& \mu(t) = (\mu-\mu_\infty)e^{-BKt}+\mu_\infty , \\
& \Sigma(t) = (\Sigma-\Sigma_\infty)e^{-2BKt}+\Sigma_\infty ,
\end{split}
\label{musigmat}
\end{equation}
where $\Sigma_\infty = W/2BK$, see  \ref{details}. In the rest of this subsection we assume the continous dynamics limit. In this limit, the expected cumulative reward picks up a divergent contribution $\propto 1/\Delta$, while the policy gradient value remains non-singular.  

It is convenient to introduce the averaged (over a normal distribution) value function $v(t,\mu,\Sigma)$, see \ref{details} for details:
\begin{multline}
v(t,\mu,\Sigma)=\mathbb{E}_{s_t\sim\mathcal{N}(\mu,\Sigma)}[V(s_t)] = \\
 -\frac{C_s+C_aK^2}{2BK}\left( (\mu-\mu_\infty)^2+\Sigma-\Sigma_\infty\right) g_2(T-t) 
-\frac{2C_s}{BK}\mu_\infty\left(\mu-\mu_\infty\right) g_1(T-t) \\ 
-\left( C_s\mu_\infty^2 + (C_s+C_aK^2)\Sigma_\infty+ \frac{C_a W}{\Delta B^2} \right) (T-t)
\label{vtmusigma}
\end{multline}
Where $g_n(\tau)= 1-\exp(-nBK\tau)$. 
Then $V(s_t)=v(t,s_t,0)$. In the steady state, as $t\to\infty$, the expected reward accumulated from time $t$ is $v(t,\mu_\infty,\Sigma_\infty)$. In that case only the linear time term (the last line) of Eq.(\ref{vtmusigma}) survives. Its three parts can be interpreted as follows. The particles fluctuate around $s=\mu_\infty$, incurring the cost $C_s\mu_\infty^2$. They experience competing effects of the diffusion (that tends to spread them) and the control (that tends to drive them to $\mu_\infty$), incurring additional costs of $(C_s+C_aK^2)\Sigma_\infty$. The last term is the divergent cost of the stochastic part of the action. The control gain $K$ only enters the second term, minimized by $K^2=C_s/C_a$, (the standard result of the optimal control theory). We will set $K$ to the optimal value in our simulations, so $\mu_\infty$ will be the only adjustable policy parameter.

The expected cumulative reward for $s_0\sim\mathcal{N}(\mu,\Sigma)$ is $v(0,\mu,\Sigma)$. Its derivative with respect to $\mu_\infty$ is the PG. Notice that, in Eq.(\ref{vtmusigma}), $\mu$ and $\Sigma$ are decoupled, so the gradient is independent of $\Sigma$. The gradient is:
\begin{equation}
\frac{\partial v(0,\mu,0)}{\partial\mu_\infty} = 
\frac{C_s+C_aK^2}{BK}\left(\mu-\mu_\infty\right) g_2(T)
-\frac{2C_s}{BK}\left(\mu-2\mu_\infty\right) g_1(T)
-2C_s\mu_\infty T
\label{polgrad}
\end{equation}

Note that at finite $\Delta$ the above expressions for the averaged value functions and the policy gradient are approximations. 

\subsection{Comparison of variance reduction methods}
\label{comparison}

We define the following approximator $\tilde Q(s_t,a_t)$: 
\begin{equation}
  \tilde Q(s_t,a_t) = r(s_t,a_t) + v(t+\Delta, s_t+\Delta B a_t, 0)
\end{equation}
The last term above is $V(s_{t+\Delta}=s_t+\Delta B a_t) + \mathcal{O}(\Delta)$. So $\tilde Q(s_t,a_t) = Q(s_t,a_t) + \mathcal{O}(\Delta)$; it becomes exact only in the limit $\Delta \to 0$. We compute $\bar V(s_t)$ from $\tilde Q(s_t,a_t)$ as defined in Eq.(\ref{vbarmf}):
\begin{multline}
\bar V(s_t) = -\Delta\left(C_ss_t^2+C_aK^2(s_t-\mu_\infty)^2\right) - \frac{C_a W}{B^2} + v(t+\Delta,s_t-\Delta BK(s_t-\mu_\infty),\Delta W)
\label{vbardiff}
\end{multline}
Note that $\bar V(s_t) = V(s_t) + \mathcal{O}(\Delta)$, while the relation between $\bar V$ and $\tilde Q$ is exact, as required in VE method. It is also straightforward to compute $\nabla\bar V(s_t)$, see \ref{details} for details:
\begin{multline}
\nabla\bar V(s_t) = 
 -\Delta\left((s_t-\mu_\infty)\left(\left(C_s+C_aK^2\right)(1-\Delta BK) g_2(T-t-\Delta)-2C_aK^2\right)\right. \\
 + \left. 2C_s\mu_\infty g_1(T-t-\Delta) \right)
\label{gradvbar}
\end{multline}

We compare the following five methods of PG evaluation:

1) No baseline (NB). PG without any variance reduction.

2) Vanilla baseline (VB). A state-independent baseline is subtracted from the cumulative reward, as in REINFORCE \cite{williams1992simple}. We use $v(t,\mu(t),\Sigma(t))$ -- approximately the value function $V(s_t)$ averaged over the distribution of $s_t$ -- as the baseline. In the steady state it becomes $v(t,\mu_\infty,\Sigma_\infty)$.

3) State-dependent baseline (SB). We use $v(t,s_t,0)$ -- approximately the value function $V(s_t)$ -- as the baseline.

4) State-action-dependent baseline (AB). We use $\tilde Q(s_t,a_t)$ as the baseline. This action dependent baseline also requires the term $\nabla\bar V(s_t)$, as in Eq.(\ref{vegrad}). This is similar to Q-Prop \cite{gu2016q}.

5) VE method, as given in Eq.(\ref{vegrad}), where the estimator $\hat Q$ is computed from Eq.(\ref{qestrecur}). Note that if we were to use the sum of rewards in place of $\hat Q$, we would recover the state-action dependent baseline method.

\begin{figure}
\includegraphics[width=\textwidth]{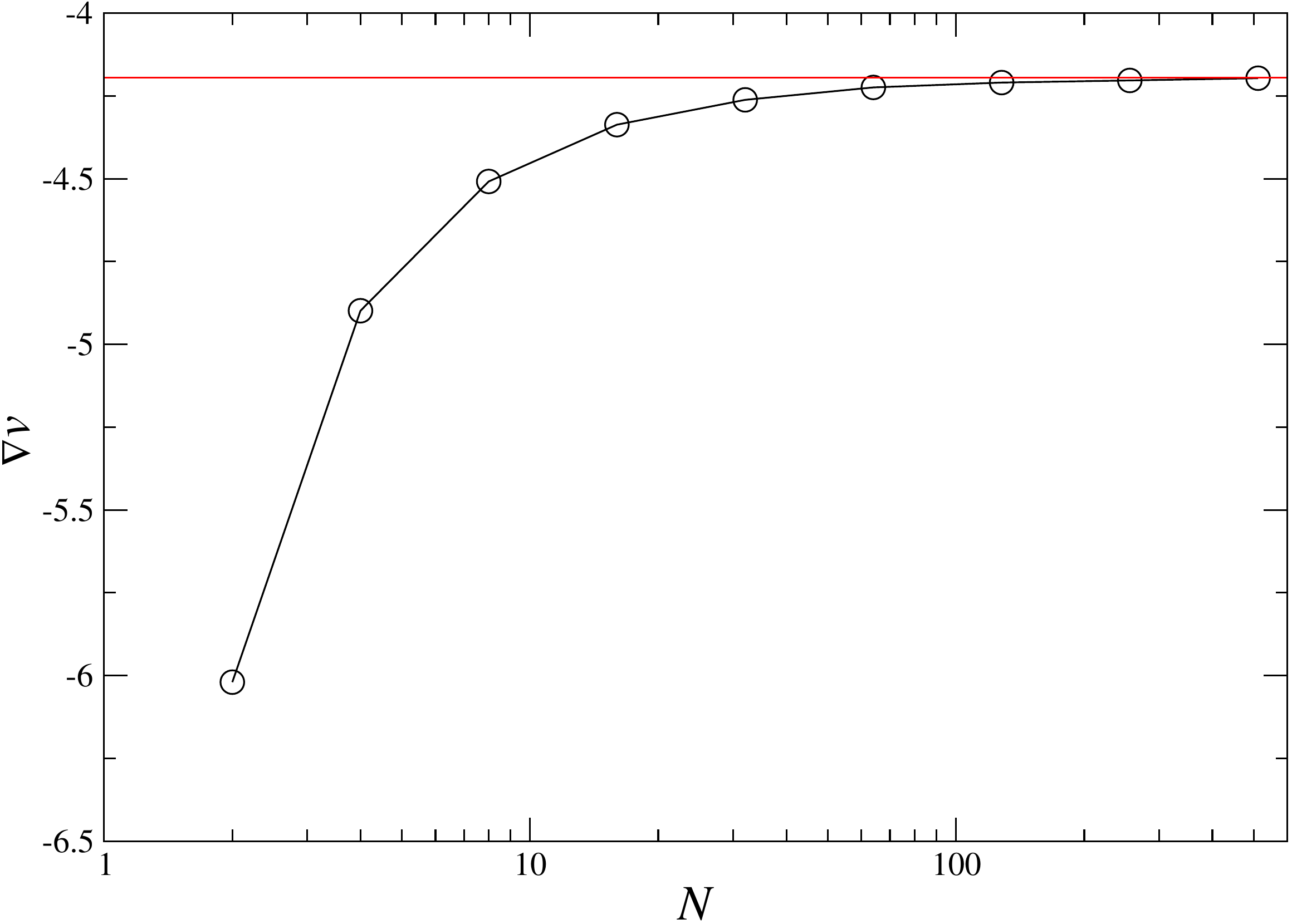}
\caption{Policy gradient $\nabla v$ for $\mu_\infty=1$ computed with VE for different $N$. The horizontal line is the theoretical value (-4.19419) in the continuous time limit.}
\label{polgradplot}
\end{figure}

We compare the above methods by simulating the controlled diffusion model (in discrete time formulation) specified by Eq.(\ref{diffdyn}) and described in subsection \ref{contrdiff}. The functions derived in the continuous dynamics limit are used as approximators, that differ from the exact functions for any finite $\Delta$.

Our quantitative investigation was performed for a single set of values: $B=W=C_s=C_a=1$. The controller gain $K$ was set to the optimal value $K=\sqrt{C_s/C_a}=1$. Since $BK$ determines the Markov chain mixing time (see Eq.(\ref{musigmat})), we set $T$ to its triple value, $T=3BK=3$. 
We set the controller parameter to $\mu_\infty=1$ and the initial state to $s_0=0$.

First, we check that our implementation of VE is indeed unbiased, by evaluating the PG expectation and making sure it agrees with the other methods for a finite $\Delta$, as well as with the theoretical value for $\Delta\to0$. Indeed, we find agreement between all the methods. Importantly, the agreement at larger values of $\Delta$, where $\tilde Q$ significantly deviates from the exact $Q$, empirically confirms the absence of bias in our implementation, corroborating our main claim of unbiased variance elimination. Convergence of the gradient to the theoretical value, $\nabla v = -4.19419$, computed from Eq.(\ref{polgrad}), is shown in Fig.(\ref{polgradplot}).

\begin{figure}
\includegraphics[width=\textwidth]{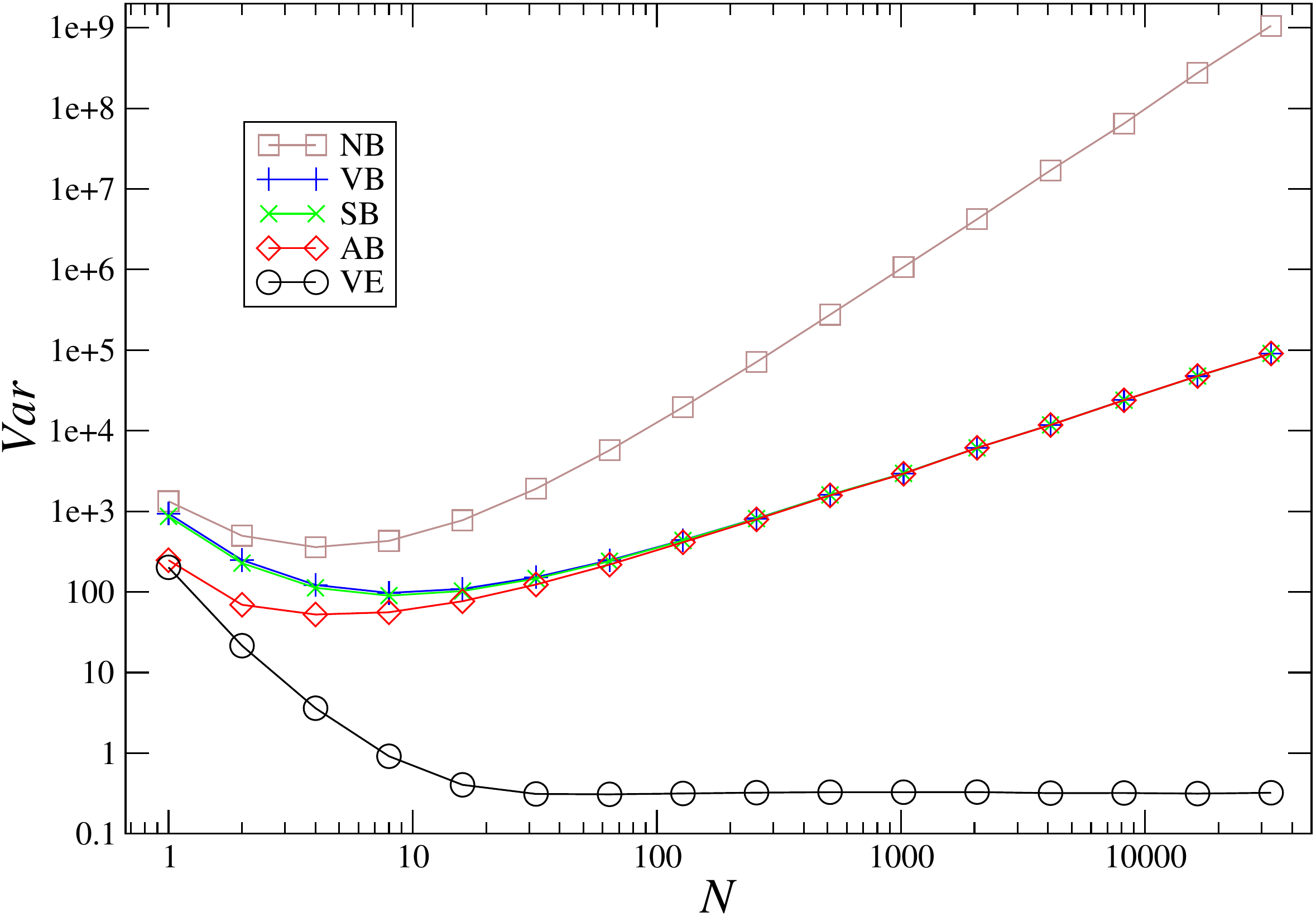}
\caption{PG variance $Var$ computed for five different methods of the gradient evaluation at different $N$. The methods, (NB, VB, SB, AB and VE), are outlined in the text.}
\label{polgradvar}
\end{figure}

In Fig.(\ref{polgradvar}) we plot the estimate of the PG variance $Var=Var(\nabla v)$ for different $N$ (with $T$ fixed). The initial drop in variance across all the methods is an artifact of the discrete dynamics at large $\Delta$, as is explained later in the section. The value function, as we defined it, has a linear time term, (see Eq.(\ref{vtmusigma})). Without a baseline, NB, it is the main source of variance, which appears to scale as $Var \propto N^2$ at large $N$. The simplest baseline, VB, that only depends on $t$, goes a long way in reducing the variance, making it $\propto N$. The state-dependent baseline, SB, brings only a mild improvement over VB, in this particular problem. Moreover, that improvement vanishes as $N$ increases: the large part of reward comes from the action cost, that scales linearly with $N$, while the state cost remains (approximately) constant. The state-action-dependent baseline, AB, does appreciably reduce variance for small N. But just as with the gain of SB over VB, the gain of AB (over SB and VB) goes away with larger $N$ as well. The limitation of AB is that it takes care of the variance due to an immediate action $a_t$, while leaving the variance originating from the rest of the trajectory mostly unchecked \cite{tucker2018mirage}. And that unchecked contribution grows linearly with $N$. This limitation is shared by all the baseline methods. We see that at large $N$, the three methods, VB, SB and AB, are almost indistinguishable in their performance. VE, on the other hand, excels at large $N$, because it takes care of all the future actions down the trajectory. Relative to the baselines, VE brings ever larger improvements as $N$ increases. Roughly speaking, in this particular model, the improvement factor over the baselines is about $10 N$ for $N>10$. Note that VE and AB merge in the limit $N\to0$, where the trajectory length is $1$. 

Notice that VE experiences a much steeper drop in variance at small $N$, than the baselines. This seems plausible, as VE's variance improves together with the quality of the approximator $\tilde Q$, which in this case improves as $N$ increases. The variance saturates at some value above $N=10$ or so. We do not investigate the nature of this value, which appears to be quite small, just about $2\%$ of $(\nabla v)^2$.

Overall, the initial drop in variance should be expected in all cases. It is an artifact of the discrete dynamics formulation. From Eq.(\ref{diffdyn}) we see that for $\Delta > \Delta_c = 2/BK$, under the discrete dynamics, particle's state diverges with time as $|s_t|\propto (\Delta/\Delta_c)^t$. The divergence of variance (among other quantities), as $\Delta$ approaches $\Delta_c$ from below, is the precursor of the impending singularity.


\subsection{Some technical details}
\label{details}

This subsection discusses some technical details omitted earlier in the section for clarity of presentation.

Consider a linear control law $\bar a = -K (s-\mu_\infty)$ and assume $s \sim \mathcal{N}(\mu,\Sigma)$. Under the linear Gaussian dynamics of Eq.(\ref{diffdyn}), after one time step, $s' \sim \mathcal{N}(\mu',\Sigma')$, where
\begin{equation}
\begin{split}
& \mu' = (1-B^dK)\mu + B^dK\mu_\infty, \\
& \Sigma' = (1-B^dK)^2\Sigma + W^d .
\end{split}
\end{equation}
Recall, that $B^d = \Delta B$ and $W^d = \Delta W$. In the continuous dynamics limit, $\Delta\to0$, the equations become:
\begin{equation}
\begin{split}
& \dot\mu = -BK(\mu - \mu_\infty), \\
& \dot\Sigma = -2BK(\Sigma - \Sigma_\infty),
\end{split}
\end{equation}
where $\Sigma_\infty = W/2BK$. Integrating these equations one obtains Eq.(\ref{musigmat}).

The expected (local) reward at time step $t$ is
\begin{multline}
\mathbb{E}_{\substack{s\sim\mathcal{N}(\mu_t,\Sigma_t) \\ a\sim\pi(a|s)}}[r(s,a)] = \\
-\left(C_s^d+C_a^d K^2\right) \left((\mu_t-\mu_\infty)^2+\Sigma_t \right)
- 2C_s^d\mu_\infty(\mu_t-\mu_\infty) - C_s^d\mu_\infty^2 -\frac{C_a^d W^d}{{B^d}^2}
\end{multline}
Taking the continuous time limit, substituting $\mu_t$ and $\Sigma_t$ from Eq.(\ref{musigmat}), and integrating from $t$ to the horizon $T$ we obtain Eq.(\ref{vtmusigma}) for $v(t,\mu,\Sigma)=\mathbb{E}_{s_t\sim\mathcal{N}(\mu,\Sigma)}[V(s_t)]$. It is easy to see that $v$ satisfies the equation
\begin{equation}
\mathbb{E}_{s\sim\mathcal{N}(\mu,\Sigma')}[v(t,s,\Sigma)] = v(t,\mu,\Sigma+\Sigma')
\end{equation}
which is useful for computing $\bar V$ in Eq.(\ref{vbardiff}).

When computing $\nabla\bar V(s_t)=\frac{\partial}{\partial\mu_\infty}\mathbb{E}_{a_t\sim\pi(a_t|s_t)}[\tilde Q(s_t,a_t)]$ one should remember to only differentiate $\pi$. Notice that the policy parameter entering $\bar V(s_t)$ in Eq.(\ref{vbardiff}) is shown explicitly as $\mu_\infty$. Therefore one way to compute $\nabla\bar V(s_t)$ is:
\begin{equation}
\nabla\bar V(s_t) = 
-\Delta C_aK^2\frac{\partial}{\partial\mu_\infty}(s_t-\mu_\infty)^2 + \left.\Delta BK\frac{\partial}{\partial\mu}v(t+\Delta,\mu,\Delta W)\right|_{\mu=s_t-\Delta BK(s_t-\mu_\infty)}
\end{equation}
from where Eq.(\ref{gradvbar}) readily follows.

\section{discussion}
\label{discussion}

Our goal in this theoretical paper was to present a novel method of variance elimination in sequential decision processes, 
thoroughly testing our findings on a simple LQG model. The possibility of a complete or nearly complete variance elimination in PG methods appears very intriguing to us. The next logical step should be to test this approach on standard benchmark problems to see how much speedup in policy learning it can offer. Our initial results on the controlled diffusion problem are very encouraging: VE outperforms the baseline methods by a factor of $10N$ in variance reduction.

What obstacles can one expect to successful practical application of VE? The method is only as good as the approximation to the value functions (and dynamics, in model-based setting). This, actually, looks promising: the bottleneck moves from the limitations of a variance reduction approach -- as is the case with the baselines -- to the quality of the approximators. Investment in the quality of approximator can now, hopefully, translate into larger steps of policy improvement and speed of convergence to an optimal solution, in the context of PG methods. Hopefully, VE can increase sample efficiency of PG, rendering it more suitable for real world learning. 

It is our impression that the current PG methods are not powerful enough to work with deeper neural net controllers. One might arrive at this conclusion by visually inspecting, for example, RL learned locomotion policies of complex 3D models, such as a humanoid robot. The energy efficiency of these policies often does not appear to be close to what one might have expected from an optimal controller. Hopefully, VE can help in attaining more economical policies in simulated, and eventually physical world tasks.

\section{Acknowledgements}

We thank Georges Harik for useful discussions. We thank Leela Hebbar for help with the manuscript.

\bibliographystyle{plain}

\bibliography{myrefs}

\end{document}